\documentclass[10pt,journal,compsoc]{IEEEtran}
\ifCLASSOPTIONcompsoc
\usepackage[nocompress]{cite}
\else
\usepackage{cite}
\fi
\ifCLASSINFOpdf
\else
\fi
\usepackage{multicol} 
\usepackage{multirow} 
\usepackage{mwe}
\usepackage{amsmath}
\usepackage[T1]{fontenc}
\usepackage[utf8]{inputenc} 
\usepackage{diagbox}
\usepackage{graphicx}
\usepackage{xcolor}
\newcommand{\tabincell}[2]{\begin{tabular}{@{}#1@{}}#2\end{tabular}} 
\usepackage{graphicx} 
\usepackage{float} 
\usepackage{subfigure} 

\begin{document}
	%
	\title{Spatial-temporal Conv-sequence Learning with Accident Encoding for Traffic Flow Prediction}
	%
	%
	%
	%
	
	\author{Zichuan~Liu, 
		Rui~Zhang,~\IEEEmembership{Member,~IEEE, }
		Chen~Wang,~\IEEEmembership{Senior Member,~IEEE, }\\
		Zhu~Xiao,~\IEEEmembership{Senior Member,~IEEE, } and~Hongbo~Jiang,~\IEEEmembership{Senior Member,~IEEE}
		
		\IEEEcompsocitemizethanks{\IEEEcompsocthanksitem Z. Liu and R. Zhang are with the Department
			of Computer Science and Technology, Wuhan University of Technology, Wuhan, China, 430070.\protect\\
			E-mail: lzc775269512@gmail.com; zhangrui@whut.edu.cn.
			\IEEEcompsocthanksitem C. Wang is with Internet Technology and Engineering R\&D Center
			(ITEC), School of Electronic Information and Communications, Huazhong
			University of Science and Technology, Wuhan, China, 430074. 
			E-mail: chenwang@hust.edu.cn. 
			\IEEEcompsocthanksitem Z. Xiao and H. Jiang are with College of Computer Science
			and Electronic Engineering, Hunan University, Changsha China, 410012.
			E-mail: zhxiao@hnu.edu.cn, hongbojiang2004@gmail.com.}
		\thanks{Manuscript received X X, 202X; revised X X, 202X. This work was supported in part by the National Natural Science Foundation of China under Grants~ 52031009, U20A20181, 61872416, 62002104 and 62071192;~in part by the Humanities and Social Sciences Foundation of Ministry of Education under Grant 21YJCZH183;~in part by the Key R\&D Project of Hunan Province of China under Grant 2022GK2020; ~in part by the Science and Technology Project of Hunan Provincial Water Resources Department under Grant~XSKJ2021000-39; ~in part by the fund of Hubei Key Laboratory of Transportation Internet of Things under Grants 2018IOT004 and WHUTIOT-2019004.~\textit{(The corresponding authors of this paper are R.~Zhang and Z.~Xiao.)}}}
	%
	%

	\markboth{IEEE Transactions on Network Science and Engineering,~Vol.~X, No.~X, MONTH~202X}%
	{Shell \MakeLowercase{\textit{et al.}}: Bare Demo of IEEEtran.cls for Computer Society Journals}
	%



	\IEEEtitleabstractindextext{%
		\begin{abstract}
			In an intelligent transportation system, the key problem of traffic forecasting is how to extract periodic temporal dependencies and complex spatial correlations. Current state-of-the-art methods for predicting traffic flow are based on graph architectures and sequence learning models, but they do not fully exploit dynamic spatial-temporal information in the traffic system. Specifically, the temporal dependencies 
			in the short-range are diluted by recurrent neural networks. Moreover, local spatial information is also ignored by existing sequence models, because their convolution operation uses global average pooling. Besides, accidents may occur during object transition, which will cause congestion in the real world and further decrease prediction accuracy. To overcome these challenges, we propose Spatial-Temporal Conv-sequence Learning (STCL), where a focused temporal block uses unidirectional convolution to capture short-term periodic temporal dependencies effectively, and a spatial-temporal fusion module is responsible for extracting dependencies of interactions and decreasing the feature dimensions. Moreover, as the accidents features have an impact on local traffic congestion, we employ position encoding to detect anomalies in complex traffic situations. We have conducted a large number of experiments on real-world tasks and verified the effectiveness of our proposed method.
		\end{abstract}
		
		}

	\maketitle

	\IEEEdisplaynontitleabstractindextext

	%
	\IEEEpeerreviewmaketitle

	\IEEEraisesectionheading{\section{Introduction}\label{sec:introduction}}

	%
	%
	%
	%
	\IEEEPARstart{W}{ith} the widespread deployment of intelligent transportation systems (ITSs), traffic handled by computational intelligence has gained more and more attention on the road to urbanization. Predicting the traffic volume in any area of the city has become one of the most fundamental problems in today's intelligent transportation system. Through identifying spatial-temporal patterns from historical data, our target is to predict the future time traffic flow of a certain road or intersection. In order to aid transportation agencies, such as traffic-police and Uber, in directing and navigating vehicles, the key problem is to build accurate volume forecasting model so that we can pre-allocate resources to avoid unnecessary blockage.
	
	\begin{figure}
		\includegraphics[width=.5\textwidth]{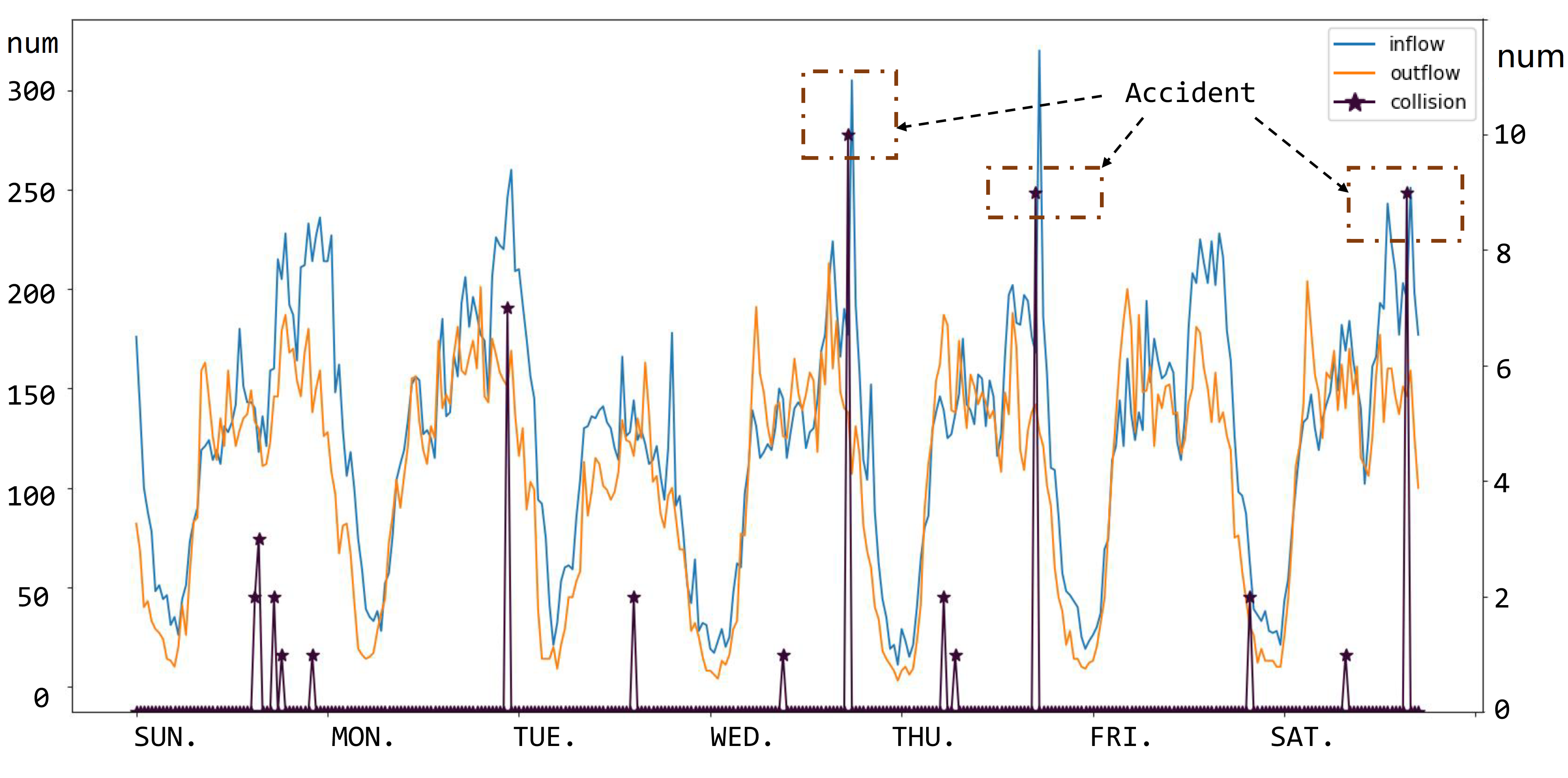}
		\caption{The impact of accidents on traffic flow. Dividing a city into grid maps by longitude and latitude, we show a week’s traffic volume in the region (x = 4, y = 9) and the accident data in this area. Traffic flow represents the number of times that vehicles enter and exit this area, and the asterisk represents the hourly number of accidents in this area. From the figure, we can observe an obvious impact of frequent accidents on traffic congestion.}
		\label{fig2}
	\end{figure}

	Focusing on utilizing real-world transportation data, some studies incorporated regional similarity features as spatial information \cite{zheng2013time, deng2016latent} and added real-world contextual data like venue, weather and events to conduct prediction \cite{pan2012utilizing, zhang2016dnn}. These studies have shown that the prediction can be improved by taking various additional factors into account. Moreover, we observe that accidents have a big impact on congestion in a certain area. Figure \ref{fig2} shows the fluctuation of traffic flow and the number of accidents in a certain week. It can be seen that the accident data increases with travel peak, which implies that a certain relationship between the two phenomena can be mined. Because it is difficult for traditional machine learning methods \cite{li2012prediction, moreira2013predicting, chen2018spatial} to capture high-dimensional spatial-temporal features, they are also unable to mine the mutations caused by traffic accidents. This observation motivates us to rethink the traffic prediction problem based on deep architecture networks.

	
	In the past few years, the deep neural network has achieved notable successes in predicting urban traffic flow by fitting the traffic volume in a certain period. Huang et al. \cite{huang2014deep} utilized a Deep Belief Network (DBN) with multitask learning to mine the considerable features and Lv et al. \cite{lv2014traffic} employed Stacked Autoencoder for forecasting traffic flow. After that, with Long Short-term Memory networks (LSTM) being widely used in the field of time series, some researchers \cite{cui2018deep, ma2015long, song2016deeptransport} used it to mine the relationship of traffic temporal correlations. Chen et al. \cite{chen2021bayesian} also enhanced the dynamics of time by Bayesian temporal factorization. These traditional deep learning methods can achieve a relatively satisfactory result. Nevertheless, all of them are mainly aimed at modeling a single sequence that only considers the time-series dependency of the transportation network without using spatial information. Besides, as recursion progresses, the temporal dependency of the short-term will be diluted to cause error propagation of surrounding information.

	Recently, researchers have applied Convolutional Neural Networks (CNN) and LSTM to traffic prediction tasks in order to exploit spatial-temporal features. Zhang et al. \cite{zhang2017deep} suggested that the traffic volume in a city during a period be regarded as pixel values. Given a set of historical traffic images, the traffic images of the next timestamp are predicted by the model, and CNN is used to model the complex spatial correlation. Yu et al. \cite{yu2017deep} proposed to apply convolution manipulation to traffic flow and the results are extracted by the LSTM network with the auto-encoder structure to capture sequential dependencies in different periods. However, spatial information is easily discarded by the global average pooling layer in CNN, and the short-term temporal dependency is also distorted and cannot receive continuous attention through the long-term dependency of the recursive network. 
	Moreover, the inefficient recurrent structure like RNN is difficult to predict large-scale traffic flow. Some researchers started to apply the Self-Attention mechanism \cite{vaswani2017attention} to capture the entire traffic flow information via scaled dot-product attention. They proposed some spatial-temporal self-attention models \cite{lin2019spatial, zhang2019self, lin2020interpretable} to deal with complex and dynamic feature dependencies simultaneously, which seems to be the most novel spatial-temporal prediction scheme. 
	Nevertheless, all of the structures asynchronously handle the dependency relationship, which ignores complex interactions and inter-prediction correlations. Besides, these methods are based on an ambiguous assumption that there are many congestions and accidents all the time, which is quite different from complex real-world traffic. According to the research of Pan et al. \cite{pan2019urban, pan2020spatio},we find that when dealing with the temporal and spatial correlation, the impact of sudden traffic from one location to another will quickly spread to their surroundings, and the state of a place at a certain moment will affect the state of the subsequent time. As shown in Figure \ref{fig2222} (a), once an accident occurs, the local state will remain for some time and quickly spread to the surrounding areas.
	
	In order to solve these drawbacks, we propose a novel model structure called Spatial-Temporal Conv-sequence Learning (STCL) to adaptively exploit spatial-temporal dependencies and consider diverse factors that affect the evolution of traffic flow in an encoder-decoder structure. In our network, we assemble the causal convolution \cite{Oord2016WaveNet} into a novel module called Focused Temporal block (FT-block), which focuses on capturing temporal dependencies in a short-range and make forward predictions with high performance. Then, the Spatial-Temporal Fusion Module (STFM in short) is applied to extract complex interactions from spatial-temporal information. Moreover, by considering real traffic conditions, we incorporate accident data to the corresponding locations via position encoding in Figure \ref{fig2222} (b). The contributions in our paper are summarized as follows: 
	
	\begin{figure}
		\includegraphics[width=.5\textwidth]{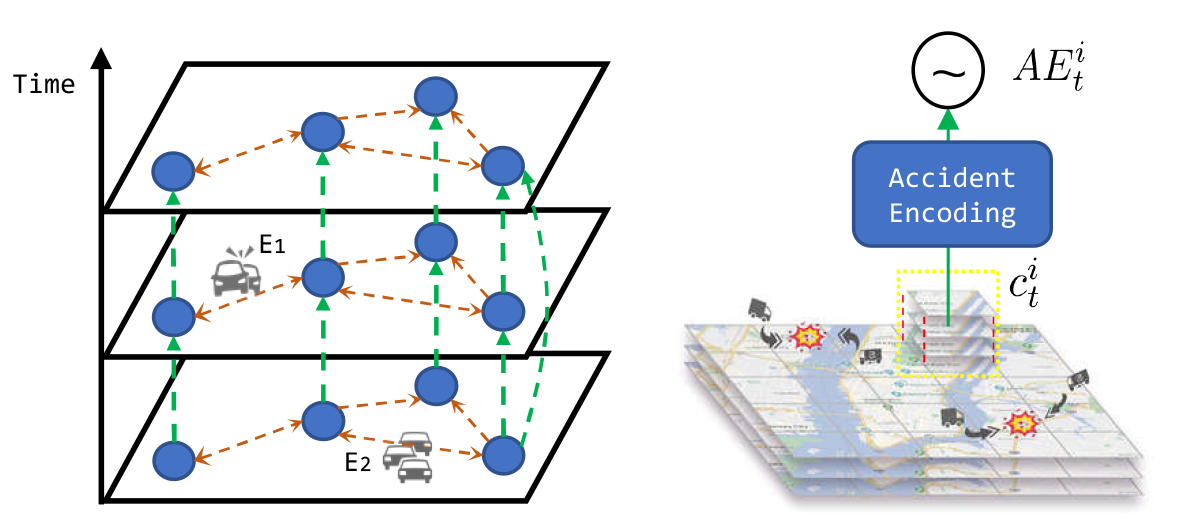}
		\caption{(a) Example of ST correlations. The brown lines show the impact of traffic between locations. For instance, in case of a car accident, the traffic congestion could quickly spread to the surrounding areas. And the green line shows the influence on the time dimension. An event in a place changes the state at a certain time and affects the state of subsequent time. (b) The accident encoding we proposed is a position encoding method. We combine accidents features $C^i$ as local traffic semantic information and merge accident data into corresponding locations via position encoding.}
		\label{fig2222}
	\end{figure}

	\begin{itemize} 
		\item In STCL, we replace the global average pooling with the newly proposed STFM, which can extract the dependencies of interactions and decrease feature dimensions. Simultaneously, we propose the FT-block to focus on capturing short-range temporal information with high-performance forward prediction.
		
		\item Since we observe the impact of accident features on local traffic congestion, we employ position encoding to detect anomalies in complex traffic situations.
		
		\item We use three baselines to compare our model against six different methods based on the neural network on large-scale real-world datasets. The results of extensive experiments indicate that our model can significantly outperform other state-of-the-art approaches. 
	\end{itemize}

	\section{Related Work}
	\label{sec:relatedwork}
	The problem of data-driven traffic prediction has received wide attention in recent years. In this section, we briefly discuss the related work of traffic flow prediction based on deep learning.

	\subsection{Traffic Flow Prediction}
	The methods based on deep learning have achieved remarkable progress in forecasting traffic flow in recent years. The LSTM \cite{hochreiter1997long, cui2018deep, ma2015long} based methods have demonstrated strong performance on capturing long-term temporal dependencies, and other studies have applied the convolution neural network to capture spatial correlation \cite{zhang2016deep, zhang2016dnn} or the deep residual convolutional structure \cite{he2016deep}. However, while these studies only focus on spatial correlation or temporal dependency, none of them pay close attention to both aspects simultaneously. Thereupon, Yao et al. \cite{yao2018deep} proposed a multi-view spatial-temporal network that learns the spatial-temporal trajectory similarity by local-CNN and integrating LSTM for demand prediction, and Wang et al. \cite{wang2018cross} proposed a transfer learning method by combining spatial-temporal information to overcome the scarcity of cross-city geographical data in diverse cities. In transfer learning and some multi-view methods mentioned above, divide-and-conquer solves spatial-temporal dependencies. Nevertheless, in these studies, the interaction between multiple modules is limited by the way of connection. They also ignore the impact of adjacent environmental factors on short-term prediction.

	More recently, several studies on road networks, such as the multi-graph convolution network \cite{geng2019spatiotemporal} and graph attention \cite{zhang2019spatial}, have applied the graph-based LSTM and CNN structure to encode the topological network of the road traffic and enhance spatial-temporal correlation. Similarly, the graph network \cite{wu2019graph, cui2019traffic} was proposed to improve the prediction accuracy by embedding each hidden spatial pattern into the graph, but the spatial-temporal dependencies pattern cannot be found on the link except for the local nodes. For the multi-connected road network, it does not only fail to well represent edge information, but also requires a lot of computational resources to propagate node information. 
	Other novel frameworks such as Yao et al. \cite{yao2019learning} introduced meta-learning using short-term data collection to tackle traffic flow prediction. Li et al. \cite{li2019predicting} incorporated R-GCN with LSTM and a novel path-embedding method for traffic network prediction. 
	These studies also overlook the impact of traffic congestion on traffic flow and what the domain of spatial-temporal correlation dissipates with the depth of networks.

	\subsection{Sequence-to-sequence Model}
	
	With the development of the encoder-decoder structure, the sequence-to-sequence method \cite{sutskever2014sequence} (seq2seq in short) has been followed with interest in the natural language processing field. The processed data is mapped by an encoder into potentially hidden vectors with the stacked structure, and then propagated and merged into each layer-decoder. Then, the decoder calculates the data at period $1:t$ steps with the encoder's output in a forward operation.

	In several studies, the attention mechanism \cite{bahdanau2014neural, vaswani2017attention} is also applied to calculate attention weights in seq2seq models. The attention mechanism \cite{vaswani2017attention} can help the decoder attract attention to each timestamp and assign calculation weight at each step to facilitate forward operation, so as to produce better results. On this basis, Yao et al. \cite{yao2019revisiting} proposed a standard attention model based on seq2seq with LSTM and local-CNN combined for traffic prediction. Another seq2seq model named spatial-temporal transformer was also proposed by Xu et al. \cite{xu2020spatial} to improve prediction accuracy.
	
	However, while the self-attention mechanism learns the whole temporal relations, the short-term time dependency will be dissipated in residual layers. Different from the existing seq2seq methods, our seq2seq method incorporates convolutional layers to represent the correlations of spatial-temporal dependencies. The experimental results show that our proposed method outperforms the previous models in traffic flow prediction.

	\section{Notations and Preliminaries}
	\label{sec:Preliminaries}
	In the following section, we first define some notations and then formulate the traffic flow prediction problem based on these notations.
	
	\textbf{Traffic Variable Definition:} We divide the whole city into a $x \times y$ grid with $r$ regions ($r = x \times y$), using $S=\left \{ s_1,s_2,...,s_r\right \}$ to denote each area and $T=\left \{ t_1,t_2,\cdots ,t_n \right \}$ to denote the whole time period for historical observations. For the definition of flow, we follow the previous work \cite{yao2019revisiting} to define the inflow/outflow regional traffic volume for an area as the number of trips arriving/departing. Specifically, when a vehicle or person is in $s_i$ at $t_i$ and appears in $s_j$ at $t_j$ ($s_i \neq s_j, t_i \leq t_j$), it contributes a value to each $s_i$’s
	outflow and $s_j$’s inflow. So the overall values of outflow and inflow of $s_i$ at period $t$ are denoted as $V^{i,t}_{in}$ and $V^{i,t}_{out}$ matrix. Similarly, in order to extract the transition between nodes, we denote $\Gamma_{in}^{i,j,t}$ and $\Gamma_{out}^{i,j,t}$ matrix to indicate transitions arriving and departing from $s_i$ to $s_j$ in period $t$. Then, we define $w$ to represent two features of input and output. Note that since transitions may span multiple regions and time intervals, we discard transitions that last longer than the threshold $m$ because they have little impact on the traffic prediction in the next time interval.

	\textbf{ Accident Variable Definition:} As shown in Figure \ref{fig2}, a traffic accident has a principal impact on traffic congestion within a certain area. In order to explore the potential relationship between them, we also use the historical accident data to deal with this task. Additional definition of accident data $C^{i,t}$ indicates the traffic congestion of $s_i$ at period $t$, so we have tensors $C\in \mathbf{R}^{x\times y\times T\times w}$. The collection and preprocessing of traffic variable data and accident data are described in in Section \ref{Description}. 
	
	\textbf{Problem Statement:} Given the historical volume and transition
	data $V\in \mathbf{R}^{x\times y\times T\times w}$ and $\Gamma\in \mathbf{R}^{x\times y\times x\times y\times T\times w}$, with the accident data $C$, the traffic flow prediction problem can be formulated as learning a function $f_\theta$ that maps the inputs to the predicted traffic flow $\hat{Y}$ at the next timestamp:
	\begin{gather}
	\hat{Y} = f_\theta ( V, \Gamma, C )
	\end{gather}
	where $\hat{Y}\in \mathbf{R}^{2r}$ and $\theta$ stand for learnable parameters.

	\section{Model Architecture}
	\label{sec:model}
\begin{figure}
		\includegraphics[width=.5\textwidth]{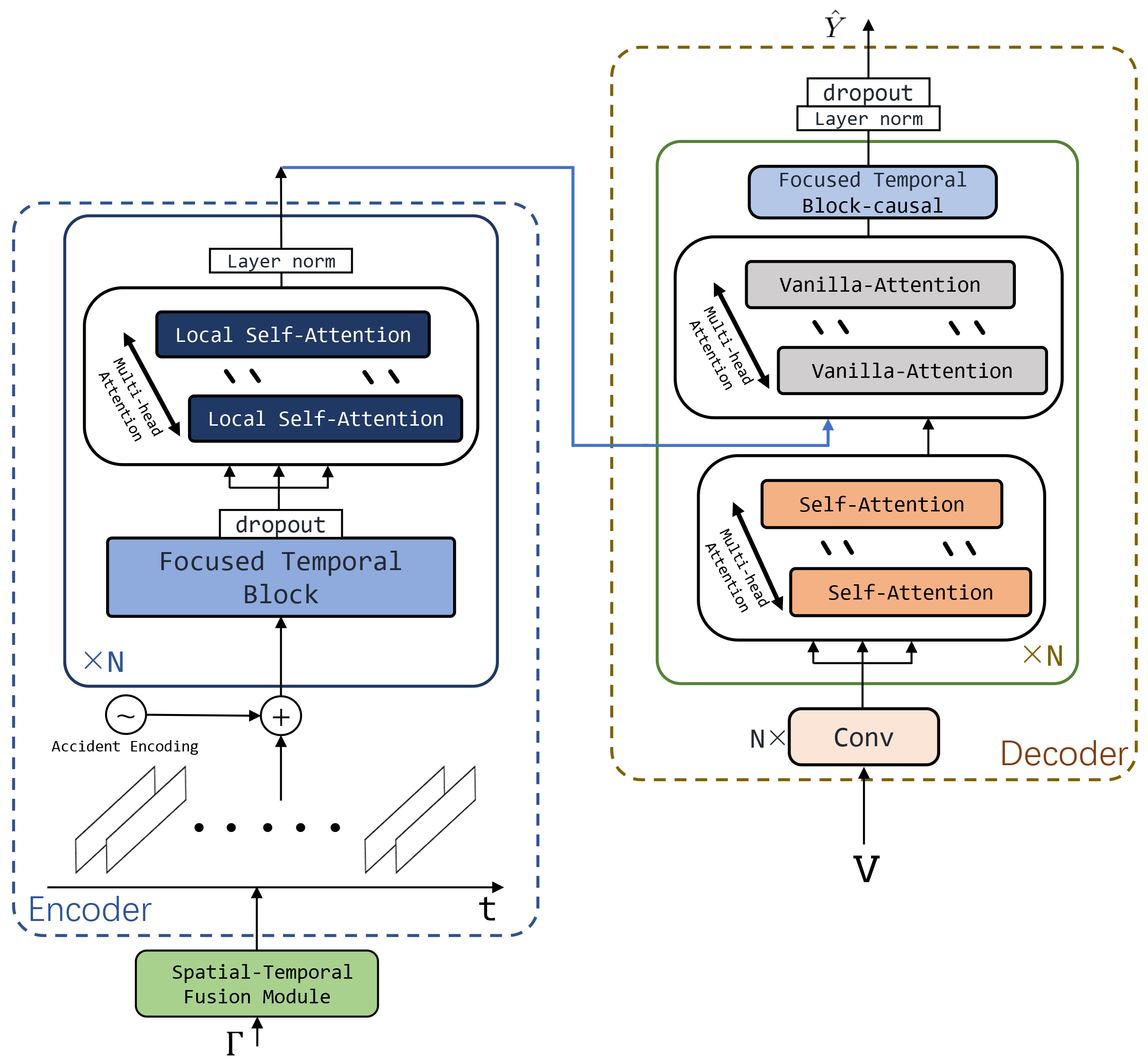}
		\caption{The architecture of our proposed Spatial-Temporal Conv-sequence Learning (STCL), which takes traffic flow and transition sequences as input and produces sequences of the volume in future time as outputs. The accident encoding module is added to the initial features of the encoder as traffic congestion, and the custom modules will be specifically introduced in the Test chapter. During training, the transition $\Gamma^{i,j,t}$ is put into the encoder for pre-training and	 the label $V^{i,t}$ at period $t$ is fed into the decoder to calculate the next data $\hat{Y}^{i,t}$.}
		\label{fig1}
	\end{figure}

	Figure \ref{fig1} shows the architecture of our proposed model, which comprehensively considers the role of spatial-temporal view. The inputs of the encoder are d-dimensional features represented by the spatial-temporal fusion module, and the accident data is projected into the encoder. Then the focused temporal block converts it into low-dimensional features and calculates it forward in a sequence-to-sequence manner. We will provide the details of our spatial-temporal fusion module, accident encoding and FT-block in the conv-seq2seq model. 

	\begin{figure*}
		\includegraphics[width=\textwidth]{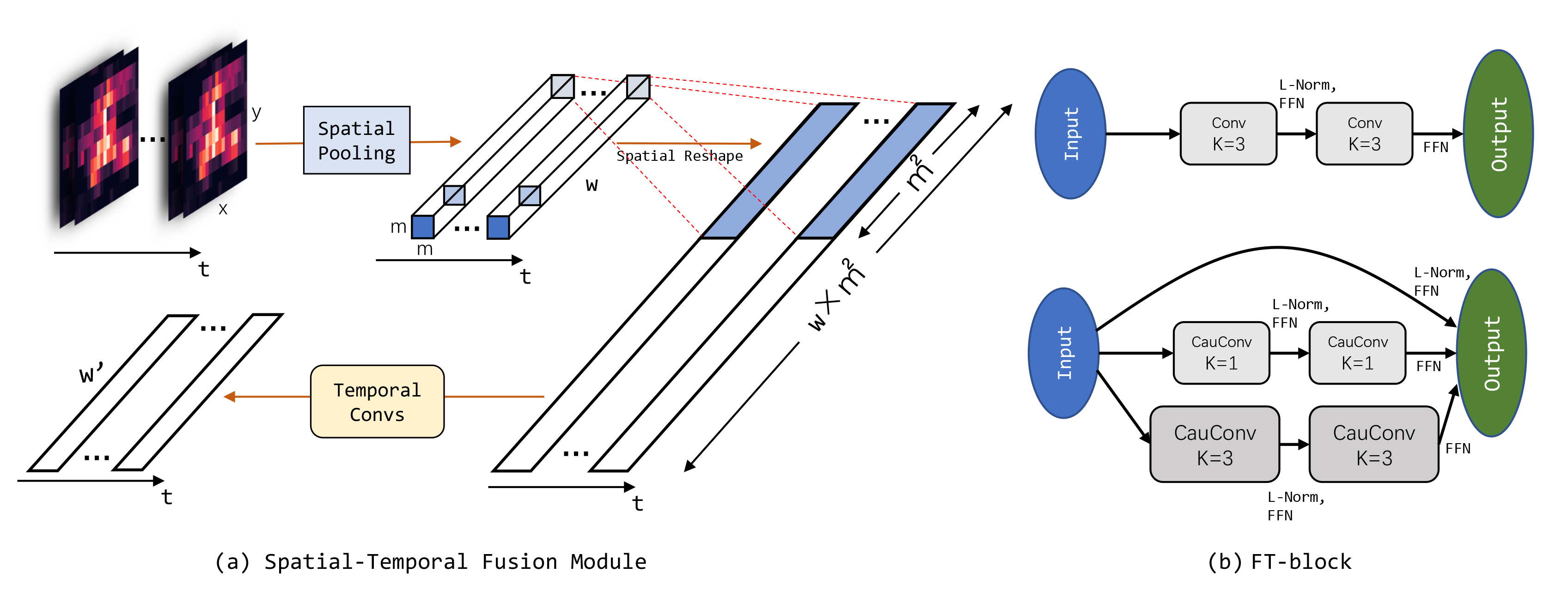}
		\caption{(a) spatial-temporal Fusion Module (STFM). The fed spatial-temporal matrix $\Gamma_{input}^{i} \in \mathbf{R}^{x\times y\times T\times w}$ is a high-dimensional feature converted to a low-dimensional feature $Z^{i} \in \mathbf{R}^{T\times w'}$. $w'$ and $w$ represent the quantity of output and input features, respectively. (b) FT-block is an incorporated implementation consisting of several branches with two same-sized causal convolution layers, where $k$ represents the kernel size of causal convolution. Each causal convolution is followed via a standardization layer and a feedforward neural network.} 
		\label{fig3}
	\end{figure*}

	%


	\subsection{Spatial-temporal Fusion Module}
	
	Before the spatial-temporal sequences are passed into the seq2seq architectures, they usually go through a stack of local-CNNs \cite{yao2018deep} to capture spatial dependencies. Since the output of CNNs cannot be directly used by the seq2seq model because of high dimensionality, most of existing methods apply global average pooling to decrease the feature dimensions. For example, the studies \cite{zhou2016learning, liu2015parsenet} showed that global average pooling is effective in dimension reduction. Nevertheless, we find that this localizing ability is not capable of continuously capturing, only indicating the attention map of CNNs and activation layers can capture subtle changes in traffic transitions. This is because the local CNN may make the same contribution to the ultimate features generated by global pooling in different spatial regions.
	
	As shown in Figure \ref{fig3} (a), we propose a Spatial-Temporal Fusion Module (STFM) that integrates high-dimensional features into low-dimensional features while retaining important local spatial information to fuse high-dimensional spatial features. 
	When generating historical transition input $\Gamma_{input}^{i}$, STFM applies a Spatial Pooling block operating across the $m \times m$ spatial dimension like RoIPooling \cite{he2015spatial}. Spatial
	Pooling extracts the domain of feature map $h_t\in \mathbf{R}^{m\times m\times w}$ by
	the setting fixed dimension from each spatial information
	$\Gamma_{input}^{i, t} \in \mathbf{R}^{x\times y\times w}$, and then reshapes the features $H^i = [h_1, h_2, \cdots, h_t]\in \mathbf{R}^{m\times m\times T\times w}$ to $M^i\in \mathbf{R}^{T\times (w*m^2)}$, where $w'$ and $w$ represent the quantity of output and input features respectively. Subsequently, to control the
	number of output channels and enhance interaction at each time interval, the $M^i$ is input into a stack of convolutions as:
	\begin{gather}
	M^{i}_k = f(M^{i}_{k-1} \ast W^{i}_k
	+ b^i_k),
	\end{gather}
	where $\ast$ represents the convolutional operation,
	$W^{i}_k$ and $b^i_k$ are two learnable parameters in the $k$th convolution layer, and $f(\cdot)$ is an
	activation function. Finally, we get the features $M'^{i}\in \mathbf{R}^{T\times w'}$ as the representation for region $i$.

	\subsection{Accident Encoding}
	As an accident has a big impact on local traffic congestion in the identified area, the centralized treatment effect can help to improve the prediction performance. To analyze the way how accidents affect traffic data, we apply map encoding to extract the features of the number of accidents and represent potential traffic congestions. Positional encoding is employed as positional accident information. In Figure \ref{fig2222} (b), the accident feature $c_t^i$ is represented as a one-hot vector $c'^i_t\in \mathbf{R}^{z+\lambda}$, where $z$ is the quantity of time intervals. We set $\lambda=7$ to represent the weekly accidents of this day and $z$ to represent the index of time interval. Therefore, the accident positional encoding of $c^i_t$ can be embedded into $AE^i \in \mathbf{R}^{T\times (z+\lambda)} $ as: 
	\begin{gather}
	AE_t^i = f(w^i_{t,1}\cdot ReLU(w^i_{t,0} c'^i_t+b^i_{t,0})+b^i_{t,1}),
	\end{gather}
	where $f(\cdot)$ is an activation function and $w^i_{t,0},b^i_{t,0}, w^i_{t,1},b^i_{t,1}$ are the learnable parameters in different layers of the network. Before feeding it into the conv-seq2seq model, we concatenate the $M'$ the whole accident encoding matrix $AE^i\in \mathbf{R}^{T\times (z+\lambda)}$ as accident features. 
	
	\subsection{Conv-seq2seq Learning}
	\textbf{Encoder-decoder: }
	Following Vaswani's work \cite{vaswani2017attention}, we employ the most competitive sequence structures of the encoder-decoder shown in Figure \ref{fig1}. The encoder maps the transition feature $M'$ processed by STFM as input. The propagated outputs with the previously predicted volume $V = 
	[v_1, v_2, ..., v_t]$ are received by the decoder to generate an output $y_{t+1}^i$ as the predicted output to represent the next time interval, where $i$ is defined as the $i$-th region in the grid.
	
	The encoder contains $N=3$ identical encoder modules that include a local self-attention module and a focused temporal block designed in Section \ref{sss}. The dimension of each encoder's outputs is set to $d_{model} = 64$. Similarly, the decoder is stacked of $N=3$ identical blocks, whose sub-layers are composed of a local self-attention, vanilla-attention module and FT-block. Where the formula of vanilla attention is similar to that of multi-head attention, it utilizes the hidden states of decoder as target features and the encoder output as source features.

	\textbf{Focused temporal block: } We follow Ma et al. \cite{ma2020forecasting} and propose the FT-block to help the map features capture short-range dynamic temporal dependencies and search their neighbors. One dimensional convolution of FT-block is defined as follows: 
	\begin{gather}
	Out^{d_o}_t = \sum_{d^i=1}^{D_i}\sum_{i=1}^{k}kernel_i^{d_id_o}*In^{d_i}_{t-k/2-1}
	\end{gather}
	where $D_i$, $D_o$ stand for the number of input and output channels, $kernel\in \mathbf{R}^{k\times D_i\times D_o}$ is the convolution kernel, $Out^{d_o}_t$ is the convolutional result learned from $In_t^{1:D_o}$ and its neighboring $k$ channels at timestamp $t$. As shown in Figure \ref{fig3} (b), the $Out^{d_o}_t$ of the employed convolution operation is not only applied to the input data at time-step $t$, but also the fusion domain features. 
	Moreover, FT-block should also be able to learn a more robust representation of traffic flow. In other words, it needs to extract dynamic ST information of traffic flow. So we link the outputs of different convolution kernel sizes to form a residual structure to prevent overfitting neighboring features with dynamic routing operations between CNNs. Then, all channels are incorporated and passed through the feedforward network to control dimension $d_{model}=64$ as the final output.
	
	Moreover, normal convolution is unsuitable for the sequence prediction because the decoder should be future-blind in a seq2seq model. Following Oord et al. \cite{Oord2016WaveNet}, causal convolution is employed to split local features into different orientations: forward and backward. In the encoder, the bidirectional convolution is performed. But in the decoder, we replace CNN layers with causal convolution in Figure \ref{fig3} (b), where only forward fusion can be performed to predict the next time. The causal convolution is defined as:
	\begin{gather}
	Out^{d_o}_t = \sum_{d^i=1}^{D_i}\sum_{i=1}^{k}kernel_i^{d_id_o}*In^{d_i}_{t-k+1}
	\end{gather}
	where $kernel\in \mathbf{R}^{k\times D_i\times D_o}$ is the convolution kernel and the feature $In_t^{1:D_o}$ is changed to forward training. Therefore, all the passed features can be blocked by unidirectional causal convolution.

	\textbf{Local self-attention: }\label{sss} Compared with ordinary language sequence models, the feature space has two additional dimensions to hold
	the domain of spatial maps, and a lot of temporal information is implicit in the overall sequence. The self-attention module \cite{vaswani2017attention} is applied to consider the entire sequence of each position to adopt the periodic temporal features under long-term dependencies. 
	
	We denote target states $Q_t = [q_1, q_2, \cdots, q_{T_t}]\in \mathbf{R}^{T_t\times d_t}$ and the keys of source states $K_s = [k_1, k_2, \cdots, k_{T_s}]\in \mathbf{R}^{T_s\times d_s}$, where $T_s$ and $ d_s$ stand for sequence length and feature dimension. The scaled dot-product
	attention is used as:
	\begin{gather}
	Att(Q_t, K_s, V_s) = softmax(\frac{W_tQ_t(W_sK_s)^T}{\sqrt{d_s}})V_s
	\end{gather}
	where $Q_t$ and $K_s$ are the same features of FT-block output, $W_t$ and $W_s$ are learnable parameters.

	Moreover, we find that due to the calculated attention weights, the model is complex for the whole input sequence and loses local temporal information for long time series in our experiments. Therefore, we propose a local self-attention block in both our encoder and decoder to capture the fixed-range dependencies as:
	\begin{gather}
	LA = softmax(\frac{W_tQ_t(W_sK_s)^T*W_m}{\sqrt{d_s}})V_s
	\end{gather}
	where $W_m \in \mathbf{R}^{T_t\times T_s} $is a mask matrix to control the range of attention, and $LA$ is the output of the local attention.
	
	Since all spatial heads can be computed in parallel, the learning and inference are computationally efficient. We adopt multi-head attention \cite{vaswani2017attention} as a function aggregating the results of all subspaces:
	\begin{gather}
	Multi\_LA = Concat(LA_1, \cdots, LA_u)W^o
	\end{gather}
	where $W^o$ is the learned linear transformation matrix and $u$ is the number of attention heads.

	\textbf{Loss function: } STCL is trained by minimizing the loss function of mean square error (MSE) between predicted result maps and truths maps:
	\begin{gather}
	\mathit{L(\theta )} = \frac{\sum_{i=1}^{r}(\hat{y^i}-y^i)^2}{r\times \omega }
	\end{gather}
	where $\hat{y^i}$ and $y^i$ represent the $i$-th traffic flow at the next timestamp respectively. $\theta$ is the learnable parameter.
	
	\textbf{Model complexity analysis: }For the original STFM and FT-block, the main time complexity is produced by convolution.Given a matrix $\Gamma_{input}^{i} \in \mathbf{R}^{ T\times w}$, the time complexity can be expressed as $O\left (Tkw^2 \right )$, where $k$ is the size of convolution kernel. The complexity of the multi-head self-attention is $O\left ( T^{2}d + Td^{2} \right )$, where $T$ and $d$ stand for sequence length and feature dimension. The complexity of the final model is determined by the number of layers and parameters.

	\begin{table}[h]
		\caption{The experimental details of the datasets.}\label{table1}
		\centering
		\fontsize{7}{12}\selectfont 
		\begin{tabular}{l|cc}
			\hline
			\hline
			\textbf{Dataset} & Taxi-NYC & Bike-NYC\\
			\hline
			\textbf{Time Span}	& \tabincell{c}{2016/01/01--2016/02/29 } & \tabincell{c}{2016/07/01--2016/08/29}\\
			\hline
			\textbf{Time Interval}& 15 minutes & 15 minutes\\ \hline
			\textbf{Grid map size} & $10 \times 20$ & $10 \times 20$ \\ \hline
			\textbf{Number of vehicles}& 24,000+ & 8,200+\\
			\hline
			\textbf{Loss threshold}& 10 & 10\\
			\hline
			\textbf{Max flow}& 928 & 168\\
			\hline
			\textbf{Total records}& 23 million & 3.2 million \\
			\hline
			\hline
		\end{tabular}
	\end{table}
	\section{Experiment}
	
	\begin{table*}
		\caption{Quantitative comparison with discriminative baselines of TaxiNYC and BikeNYC, and the results were predicted by RMSE and MAE.	}\label{main}
		\fontsize{9}{13}\selectfont 
		\centering
		\begin{tabular}{l||c|c||c|c}
			\hline
			\hline 
			\multirow{2}{*}{\textbf{\diagbox{Methods}{Datasets}}}&\multicolumn{2}{c||}{\textbf{TaxiNYC}}&\multicolumn{2}{c}{\textbf{BikeNYC}}\\\cline{2-5}
			& \textbf{Inflow}(RMSE/MAE)& \textbf{Outflow}(RMSE/MAE)&\textbf{Inflow}(RMSE/MAE)& \textbf{Outflow}(RMSE/MAE)\\
			\hline
			HA & 25.9856 / 17.2775& 23.6287 / 14.9653 & 11.1288 / 7.9088 &10.8701 / 7.8254\\
			ARIMA \cite{box2015time} & 53.2120 / 33.7210 & 46.2979 / 34.1989 & 27.4020 / 23.4436 & 27.6041 / 23.6154\\
			VAR \cite{lutkepohl2005new} &41.5135 / 26.9960& 39.4129 / 17.3546& 16.9276 / 13.7604& 16.5867 / 13.6108\\\hline
			MLP & 19.0216 / 13.1050 & 16.0133 / 11.0316 & 9.5292 / 6.8345& 9.1321 / 6.5823\\
			FC-GRU &18.6762 / 12.6428 & 16.3984 / 11.1141 & 9.4131 / 6.8256 & 9.1179 / 6.6041\\
			ConvLSTM \cite{xingjian2015convolutional} & 18.6523 / 12.6978 & 16.1923 / 11.0207 & 9.3754 / 6.6942 & 9.2677 / 6.7002\\
			ST-ResNet \cite{zhang2017deep} & 18.5457 / 11.2446 & 16.0472 / 9.8477 & 9.0788 / 6.7831 & 8.9774 / 6.6475\\
			DMVST-Net \cite{yao2018deep} &17.3786 / 10.8742 & 15.5972 / 9.1285 & 8.9796 / 6.3568 & 8.6871 / 6.1473\\
			STDN \cite{yao2019revisiting} &16.8271 / 9.7584 & \textbf{15.3281 / 8.7156}& 8.5764 / \textbf{ 5.8981} & \textbf{7.9934} / 6.1382\\
			\hline
			\textbf{STCL w/o AE(ours)} & \textbf{16.6148 / 9.4863 }& 15.4652 / 8.6812 & \textbf{ 8.3766} / 5.9184 & 8.0345 / \textbf{ 6.0918 }\\
			\textbf{STCL(ours)} & \textbf{16.2752 / 9.1295} & \textbf{14.9671 / 8.2187 }&\textbf{ 7.9184 / 5.8441 }&\textbf{ 7.7815 / 5.8991 }\\
			\hline
			\hline
		\end{tabular}
	\end{table*}

	\label{sec:experiment}
	\subsection{Datasets and Preprocessing}\label{Description}
	The proposed deep architecture model is evaluated on two large-scale real-world datasets from New York City (NYC), TaxiNYC and BikeNYC. In addition, we add NYC motor vehicle accident data to support our accident encoding.

	\textbf{TaxiNYC and BikeNYC: } 
	The New York data (NYC) we use includes a total of 60 days of traffic records in 2016. In addition, each record also includes the positions of the vehicles and the start and end times of the trip. A summary of the two datasets is shown in Table \ref{table1}, and the transition implementations and volumes preprocessing methods are similar to that of Yao et al. \cite{yao2019revisiting} which is the most commonly used approach in industry and academia. 
	
	\textbf{Accident Data: } Vehicle accident data is collected from the New York City Open Data\footnote{https://opendata.cityofnewyork.us/}, and there are about 8,000 traffic accident records in New York every month. The data is processed in the same way as above with time slot, $x$ location and $y$ location, and the data values are the number of accidents at each time point in each grid.

	\subsection{Implementation Details}
	For training strategies, although cross validation is effective in most tasks, researchers \cite{bergmeir2018note} usually worry it will lead to data leakage in time series analysis. Here we apply the previous 45 days, 75\% of the data, as the training sets and the last 15 days as the testing sets. 	Before feeding them into the network, we perform Min-Max scaling to convert the training and testing datasets to the interval from 0 to 1, and denormalize the predicted value for evaluation after the prediction. For STFM information, we set the spatial pooling size $m = 5$, the convolution kernel size $5\times 5$, and the number of output channels to $256$. In our proposed conv-seq2seq model, we set convolution kernel sizes $3\times 3$ with $64$ filters, the $d_{model} = 64$, $d_{f} = 256$, and the dropout rate is set to $dp_{rate} = 0.1$. $N$ of the model stacked and the multi-head attention modules are set to $3$ and $4$, respectively. To schedule the learning rate, we apply the so-called Noam scheme as introduced in \cite{vaswani2017attention} with warm-up steps set to 4000. We tune the STCL model on the validation set and observe that using the hyperparameters above has achieved the best performance in both data sets. It takes about six hours to experiment with 64 as the batch size on a computer with four NVIDIA RTX1080Ti GPUs.	

	
	\subsection{Methods for Comparison}
	In this subsection, we compare the proposed model with nine discriminative traffic flow forecasting approaches and analyze the results. These approaches can be divided into three categories: (1) traditional time series prediction models; (2) spatial-temporal features mining deep learning network for traffic flow prediction; and (3) spatial-temporal prediction networks with multi-view or complex structure. The details of these compared approaches are described as follows:
	
	\begin{itemize} 
		\item \textbf{HA:} Historical average is a statistical method, which can directly calculate the coming traffic flow by averaging data of the corresponding period in historical flow.

		\item \textbf{ARIMA \cite{box2015time}:} Auto-regressive integrated moving average model is universally known a statistical mathematical method that usually makes time trend prediction by moving average.
		
		\item \textbf{VAR \cite{lutkepohl2005new}:} Vector auto-regression
		is usually applied to capture interconnected time-series.

		\item \textbf{MLP:} Multi-layer perceptron with three fully connected layers. We set the numbers of hidden units with \textit{relu} activation to 128, 64 and 32, respectively.

		\item \textbf{FC-GRU:} FC-GRU is the vanilla version of GRU. It is a very classic time prediction network and outperforms LSTM in some cases. We flatten the traffic volume tensor and take it as input into FC-GRU and we set the number of the hidden units to 64.
		
		\item \textbf{ConvLSTM \cite{xingjian2015convolutional}:} Convolutional LSTM is a well-known network to forecast temporal data. We evaluate multiple hyperparameters and choose the best setting: unit = 64 and learning rate = 0.001.

		\item \textbf{ST-ResNet \cite{zhang2017deep}:} spatial-temporal Residual Convolutional Network incorporates the closeness, periodic dependency, and trend data with external weather features to forecast the citywide traffic flow with residual structure.

		\item \textbf{DMVST-Net \cite{yao2018deep}:} Deep Multi-View spatial-temporal Network, this framework makes predictions through multiple perspectives, including time modules, spatial modules, and external factors like weather.
		
		\item \textbf{STDN \cite{yao2019revisiting}:} Spatial-Temporal Dynamic Network predicts future traffic flow, which takes the long-term periodic dependency and temporal lagging simultaneously by the attention module. 
		
	\end{itemize}

	To test the performance of our proposed model, we evaluate traffic prediction errors using the Root Mean Square Error (RMSE) and Mean Absolute Error (MAE). We use Adam \cite{kingma2014adam} as the optimizer for all baselines, set the parameter of Adam $\beta_1 = 0.9$, $ \beta_2=0.98$, $\epsilon=10^{-9} $ and use the mirrored strategy for distributed training. 
	For ST-ResNet, DMVST-Net and STDN, the hyperparameters remain as the optimized settings, which is the same as their authors’ suggestion. The model training time increases as parameter quantity grows. STDN and STCL take longer time than other methods, about 5.6 hours and 6.2 hours on the taxi dataset, respectively. Since STDN has fewer learnable parameters than STCL, it is a bit faster. The time it takes to train the models is significantly longer than the time it takes to predict. After being trained, all of these methods can produce predictions in minutes.

	The testing results on two datasets of the test set are shown in Table \ref{main}. The RMSE and MAE in TaxiNYC are both higher than those of BikeNYC on the methods because the traffic volume in TaxiNYC is presented on a larger scale than BikeNYC. For the two data sets, the baseline traditional time series analysis methods suffer a large prediction deviation on two metrics. This indicates that the traffic flow is largely affected by spatial variances in consecutive time intervals, and it is insufficient to use manual feature extraction to mine temporal information. Compared with neural-network-based models like MLP, the worse results in basic methods confirm that complex spatial-temporal dependencies cannot be well captured with simple regressions. Thus, the inter-dependency among regions could be explored to achieve better performance.

	\begin{table}[h]
		\caption{Further analysis of the role of the accident module.}\label{table4}
		\centering
		\fontsize{8}{11}\selectfont 
		\begin{tabular}{c|l|cc|cc}
			\hline
			\hline

			\multirow{2}{*}{\textbf{Datasets}}&\multirow{2}{*}{\textbf{Methods}} &\multicolumn{2}{c}{\textbf{with acc}}&\multicolumn{2}{c}{\textbf{w/o acc}}\\\cline{3-6}
			& & Inflow& Outflow& Inflow& Outflow\\
			\hline
			&MLP*	& 19.17& \textbf{15.91} & 	\textbf{19.02} &16.01 \\
			&GRU*	& \textbf{18.58}&\textbf{16.15 } & 18.67 & 16.39\\
			TaxiNYC&ConvLSTM*	& 18.88 & \textbf{15.99} & \textbf{18.65} & 16.19\\
			&STDN*& \textbf{16.47} & \textbf{15.11} & 16.83 & 15.33 \\ 
			&\textbf{STCL(ours)}& \textbf{\color{red} 16.28} &\textbf{\color{red} 14.97} & 16.61& 15.46 \\
			\hline
			&MLP*	& \textbf{9.42} & 9.19 & 	9.52 &\textbf{9.13} \\
			&GRU*	& 9.44 & \textbf{9.08} & \textbf{9.41}& 9.11\\
			BikeNYC&ConvLSTM*	& \textbf{9.32} & \textbf{9.16} & 9.37& 9.26\\
			&STDN*& \textbf{8.24} & \textbf{7.78} & 8.57 & 7.99 \\ 
			&\textbf{STCL(ours)}& \textbf{\color{red} 7.92} &\textbf{\color{red} 7.78} & 8.37& 8.03 \\
			\hline
			\hline
		\end{tabular}
	\end{table}

	For neural-network-based methods, our method outperforms ConvLSTM and ST-ResNet on both metrics. The possible reason is that they fail to dig out the potential temporal sequential dependencies and neighborhood dependencies. For instance, ST-ResNet focuses on long-range input information through deep residual CNNs, which makes it inefficient in measuring the surrounding correlation. 
	Due to the pooling effect, the smaller grid information will be averaged by convolution operation, which drags down the prediction accuracy of most networks with CNNs.
	Recent work has solved this problem by combining CNNs and LSTMs. Thereupon, DMVST-Net and STDN attempt to solve the dissipation of spatial-temporal semantics through multi-view and long-term attention mechanisms, respectively. However, they are also faced with the above-mentioned problems for not taking into account local semantic changes, such as accidents. Therefore, the experiment shows that STCL achieves the best performances on RMSE and MAE, 2.65\% - 7.51\% lower than DMVST-Net and STDN. There are two main reasons: (1) The short-range temporal dependencies receive no extra attention in the dynamic spatial similarity, but STCL takes the whole traffic volume tensor as input, using the FT-block to capture the spatial-temporal information of the field and the multi-head attention mechanism to capture long-term periods. (2) Local spatial information is discarded because of the application of global average pooling in CNN, so we design STFM to capture the local spatial information.

	Note that STCL adds accident data as spatial encoding features to represent local traffic congestion, significantly outperforming the networks without accident encoding structure. As shown in Table 3, we further analyze the role of accident data in traffic flow prediction through RMSE. The symbol * represents that we concatenate accidents as features to traffic features, and the two columns on the right represent whether there is an accident module. The performance of most of the above-mentioned methods has been improved, especially the models considering spatial dependency. This is because some models, such as GRU or MLP, only focus on historical time information but cannot dig out the spatial relationship and additional local traffic semantic information from external features. 
	However, for some models that consider the spatial relationship after the convolution operation, the dependencies in the domain area are transmitted so that the traffic congestion information can be mined. Compared with other baselines, our model has been empowered significantly by position encoding. It illustrates that the accident data can extract the local semantic information varieties and apply them to the traffic flow prediction.
	
	\subsection{Hyperparameters Analysis}

	To demonstrate the impact of various hyperparameters, we also evaluate different settings to measure the corresponding performance on the TaxiNYC dataset. By adjusting our model parameters on the validation set, we observed that $d_{model} = 64$, $d_f = 256$, $dp_{rate} = 0.1$, $m_h = 4$, $N = 3$ achieve the best results in both datasets in the experiments above. It takes about seven hours for TaxiNYC and five hours for BikeNYC when batchsize = 64. In Table \ref{tablsse3}, we record the experimental results of adjusting various parameters by RMSE with the control variable method, where the blank space represents the same parameters as STCL.
	
	\begin{table}[h]
		\caption{Influence of different super parameters on the prediction results.}\label{tablsse3}
		\centering
		\fontsize{9}{12}\selectfont 
		\begin{tabular}{l|cccc|c}
			\hline
			\hline
			
			& $d_{model}$& $d_f$& $dp_{rate}$&$N$&Inflow/Outflow\\
			\hline
			\multirow{3}{*}{$d_{model}$}	& 32 & & & &17.35/16.13\\
			& 128 & & & &16.41/15.57\\
			& 256 & & & &16.33/\textbf{14.97}\\
			\hline
			\multirow{2}{*}{$d_f$}& &128& &&16.62/15.34\\
			& &512& &&\textbf{16.26}/15.02\\
			\hline	
			\multirow{2}{*}{$dp_{rate}$}& &&0.05 & &17.14/15.67\\
			& &&0.2 & &16.97/15.78\\
			\hline
			\multirow{2}{*}{	$N$}& && &2& 16.75/15.27\\ 
			& && &4&	16.34/15.01\\
			\hline
			\textbf{STCL}& 64 &256&0.1& 3& 16.28/\textbf{14.97}\\
			\hline
			\hline
		\end{tabular}
	\end{table}
	As shown in Table \ref{tablsse3}, the influence of different parameters on the model is similar. To some extent, when the model is larger, the performance is better. However, as a larger model contains more parameters, its training time increases exponentially. In other words, overfitting arises. Due to different random numbers, some swings will be generated in each experiment. In order to stabilize the model, we adopt the baseline parameters as the final experimental results to prevent overfitting.
	
	\subsection{Component Analysis}
	To better understand the impact of components on the STCL, we also make a controlled variable experiment to examine the roles of spatial-temporal fusion module and FT-block:
	
	\begin{figure}
		\includegraphics[width=.5\textwidth,height=.31\textheight]{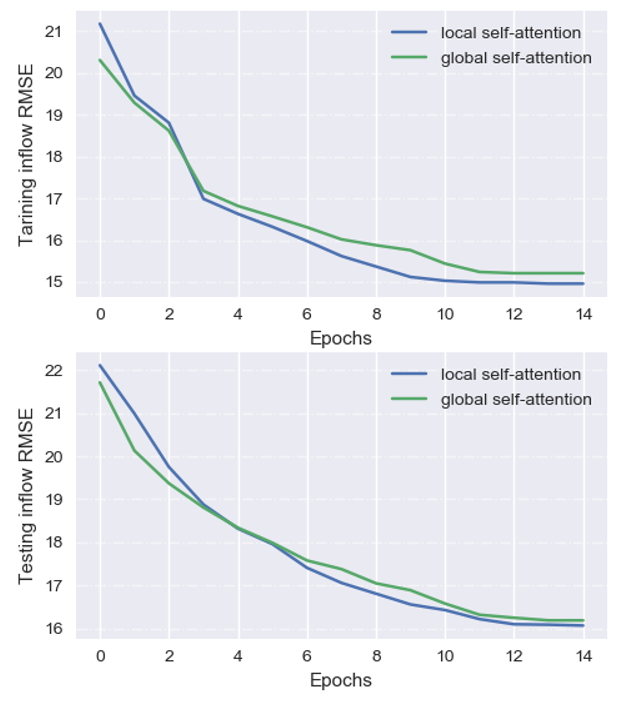}
		\caption{Loss reduction of local and global attention mechanisms. The local self-attention mechanism is slightly superior to the global one in terms of traffic inflow. The two methods exhibit little difference in the prediction of the training set and testing set, which shows that local self-attention can achieve a similar effect as the global one through short-term information.}
		\label{figsssss}
	\end{figure}

	\begin{itemize} 
		\item \textbf{multi-head attention:} For this variant, we remove all redundant structures, only just inputting the data into the multi attention mechanism and decoding the sequence by using the total concatenation layer.
		\item \textbf{w/o STFM:} For this variant, we only remove the STFM module and feed the transition features $\Gamma^{i}$ into Conv-seq2seq Learning after processing.
		\item \textbf{local CNN: } We replace the module with local CNN that is proposed by Zhang et al \cite{zhang2017deep} to extract local spatial features. 
		\item \textbf{w/o FT-block:} Feeding the extracted periodic features directly into the multi-head attention mechanism.
		\item \textbf{FT-block-CNN:} The module in the Decoder uses an ordinary CNN to replace the causal convolution of forward prediction in the last step.
		\item \textbf{global self-attention:} Compared to local self-attention, we use global self-attention instead, and other structures are consistent with STCL.
		
	\end{itemize}

	The impact of the above-mentioned structures on the reliability of STCL is shown in Table \ref{table3}. Among them, some cases are meaningless when the multi-head attention mechanism only gives the weight position of data and does not represent major features. For other variants, we first compare the model without STFM and then replace the module with local CNN \cite{zhang2017deep}. The results show that STFM is slightly better than local CNN in extracting spatial dependencies. We can infer that it is important to have a dynamic spatial module feature extraction to predict traffic flow. This is because, without the effect of the pooling layer, STFM can better capture the dynamic spatial dependencies. Similarly, we experiment by removing the FT-block and replacing the causal convolution in the FT-block with bidirectional convolution layers (CNN layers). The results show that STCL is 1.59\%-7.70\% lower than and the case when FT-block is removed. Because after we add the causal convolution for forwarding prediction, the module will pay more attention to the traffic at the next moment rather than the global temporal.

	\begin{table}[h]
		\caption{Analysis of some components of STCL.}\label{table3}
		\centering
		\fontsize{8}{12}\selectfont 
		\begin{tabular}{l|cc|cc}
			\hline
			\hline
			\multirow{2}{*}{\textbf{\diagbox{Methods}{Datasets}}} & \multicolumn{2}{c}{\textbf{TaxiNYC}}&\multicolumn{2}{c}{\textbf{BikeNYC}}\\\cline{2-5}
			& Inflow& Outflow& Inflow& Outflow\\
			\hline
			multi-head attention	& 31.51 & 27.34 & 23.52 & 20.48\\
			w/o STFM	& 18.47 & 17.13 & 10.52 & 9.47\\
			local CNN	& 17.15 &15.24&8.61& 8.28\\
			w/o FT-block& 17.41 & 15.57 &8.84&8.21\\ 
			FT-block-CNN &16.68 & 15.03 & 8.16 &7.99\\ 
			global self-attention &16.29 & \textbf{ 14.95} & \textbf{7.91} &7.83\\ 
			\hline
			\textbf{STCL(ours)}& \textbf{16.28} &14.97& 7.92& \textbf{7.78} \\
			\hline
			\hline
		\end{tabular}
	\end{table}

		\begin{figure}
		\includegraphics[width=.5\textwidth]{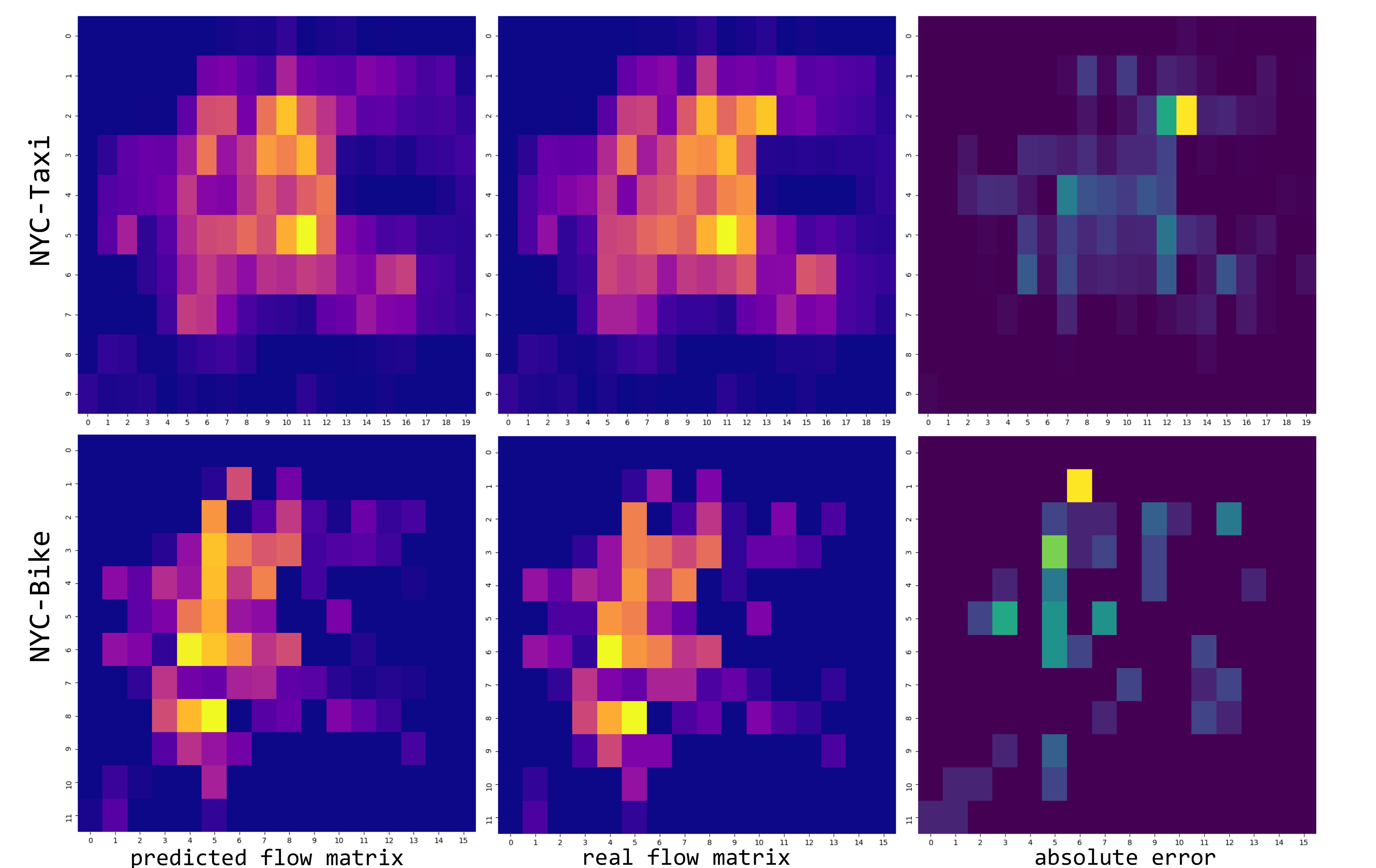}
		\caption{Flow matrix at the latest timestamp on two datasets. Comparison between the truth inflow/outflow matrix and the predicted flow of the latest timestamp. On the right is the residual matrix of subtracting two matrices, where the prediction accuracy in different regions can be seen.}
		\label{figs}
	\end{figure}
	
	The width of attention in the sequence model has a great influence on the distribution of attention weights. Therefore, we do an experiment to train a Conv-seq2seq model with global multi-headed attention, whose input sequence and other parts of the model are the same as those of SCTL. During training, we replace the weights beyond the selection range with zero and only keep the required weights to ensure the effective width of attention weights. Experimental results in Figure \ref{figsssss} show that local self-attention is better than the global one. Although the initial loss of local self-attention is higher, its loss decreases quickly and its running time is about 40 minutes faster than the global one in TaxiNYC. Moreover, in test data, the performance of local self-attention has been improved slightly. This illustrates that the local attention mechanism is effective in the prediction of short-term dynamic dependencies, while long-term time series may variate and appear non-periodically. Eventually, the best performance is achieved when all components are combined. As shown in Figure \ref{figs}, high prediction accuracy is achieved in most regions, indicating that this model has a high performance in traffic flow prediction.


	\subsection{Case Study}
	
	In this section, we will illustrate our STCL in handling congestion, slight congestion, and smooth-traffic area. Then, we will examine the role of Accident Encoding in traffic prediction.

	\begin{figure}
		\centering 
		\subfigbottomskip=2pt 
		\subfigcapskip=-5pt 
		\subfigure[Congested region (x = 5, y = 11)]{
			\includegraphics[width=\linewidth]{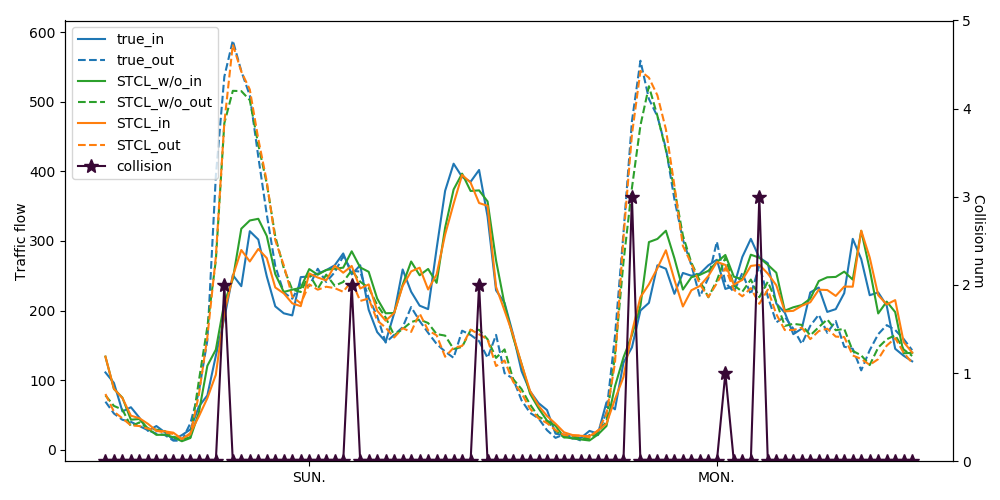}}
		\\
		\subfigure[Slightly crowded region (x = 4, y = 7)]{
			\includegraphics[width=\linewidth]{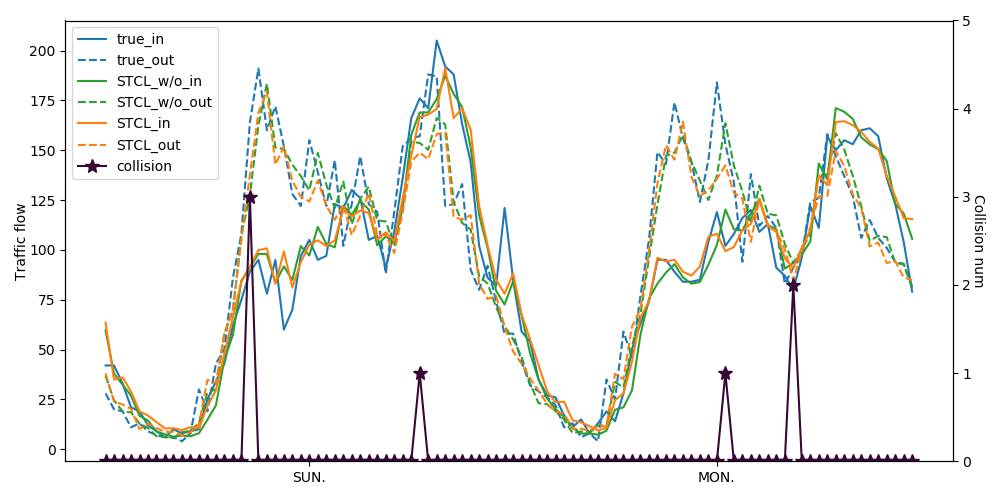}}
		\\
		\subfigure[Smooth region (x = 7, y = 4)]{
			\includegraphics[width=\linewidth]{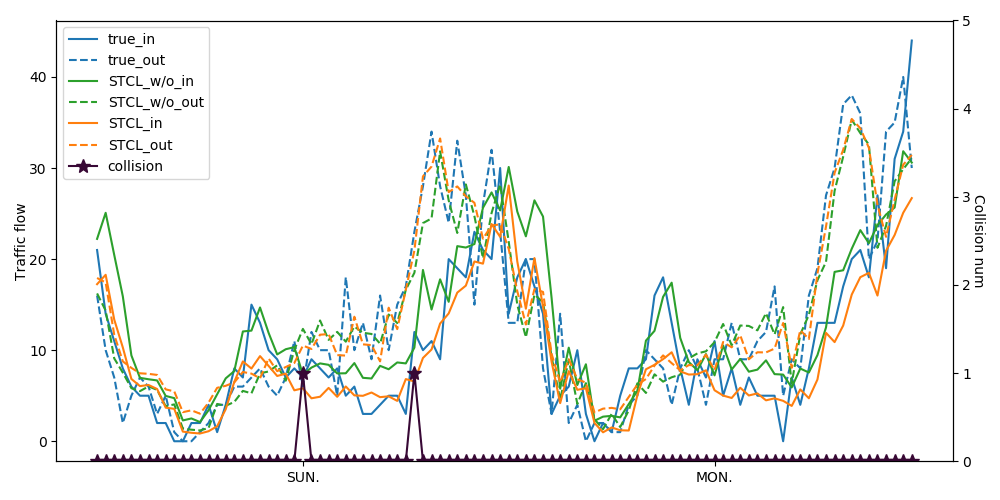}}
		\caption{The fitting effectiveness of our model on different traffic areas.}		\label{adadasdasd}
	\end{figure}

	According to traffic flow theory, traffic flow would be relatively low when either traffic congestion occurs or there is little traffic demand on roads. When congestion occurs, the outflow will be significantly higher than the inflow. However, when there is less demand, the values of inflow and outflow will be relatively low and quite close to each other. For an illustration of our proposed STCL, we compare our predicted results with true values. Specifically, we select three different areas from the test dataset to represent three traffic conditions: congestion, slight congestion, and smooth-traffic area. We use colors to distinguish different methods in Figure \ref{adadasdasd}, where solid lines denote inflow and dotted lines represent outflow; x and y represent our selected region.

	%
	%
	%
	As can be seen from the figures, the congested area and the slight congestion area show a similar trend: the traffic flow gradually decreases in 0 a.m.-5 a.m. period, and then gradually rises to the peak value in the morning with constant fluctuation. The traffic volume reaches the evening peak around 6 p.m., and then gradually decreases. Considering morning and evening peaks, the inflow will be significantly lower than the outflow when there is special congestion. When the traffic flow decreases, the inflow will start to grow. But in the period of small traffic flow, both of the two scenarios above show a similar trend. For the smooth-traffic area, the flow fluctuates gently after the first peak of the day. Therefore, judging from the value of inflow and outflow, it is easy to distinguish between a traffic jam and little road traffic demand.

	To examine the role of Accident Encoding, we have made a further comparison in the same conditions by activating and deactivating Accident Encoding. When no special event occurs, the curves of the two methods look similar. But when a special event occurs, as shown in Figure \ref{adadasdasd}, Accident Encoding will play a significant role in case of congestion: the curves of the predicted inflow and outflow will obviously fit the true value. In low-traffic areas, the probability of events is also relatively low, and the predicted value fluctuates.


	\section{Conclusion and Outlook}
	\label{sec:outlook}
	In our work, spatial-temporal conv-sequence learning is proposed to improve short-term temporal dependencies by FT-block and sufficiently extract the utilization of domain spatial-temporal information via STFM. In particular, we have introduced traffic accident data as input for accident encoding to explore the impact on regional traffic congestions. The results indicate that our proposed method has superior performance of prediction on both TaxiNYC and BikeNYC datasets compared with other state-of-the-art models. But the potential problem is that our model will cause noise and error propagation in long-term prediction. In the future, we may use semantic track information with 	the Point of Interest (POI) in each area and achieve accurate 	traffic prediction by retrieving nontrivial information.
	\ifCLASSOPTIONcaptionsoff
	\newpage
	\fi

	
	
	\bibliographystyle{IEEEtran}
	\bibliography{paper}
	%
	
	%
	%
	
	%
	
	\begin{IEEEbiography}[{\includegraphics[width=1in,height=1.25in,clip,keepaspectratio]{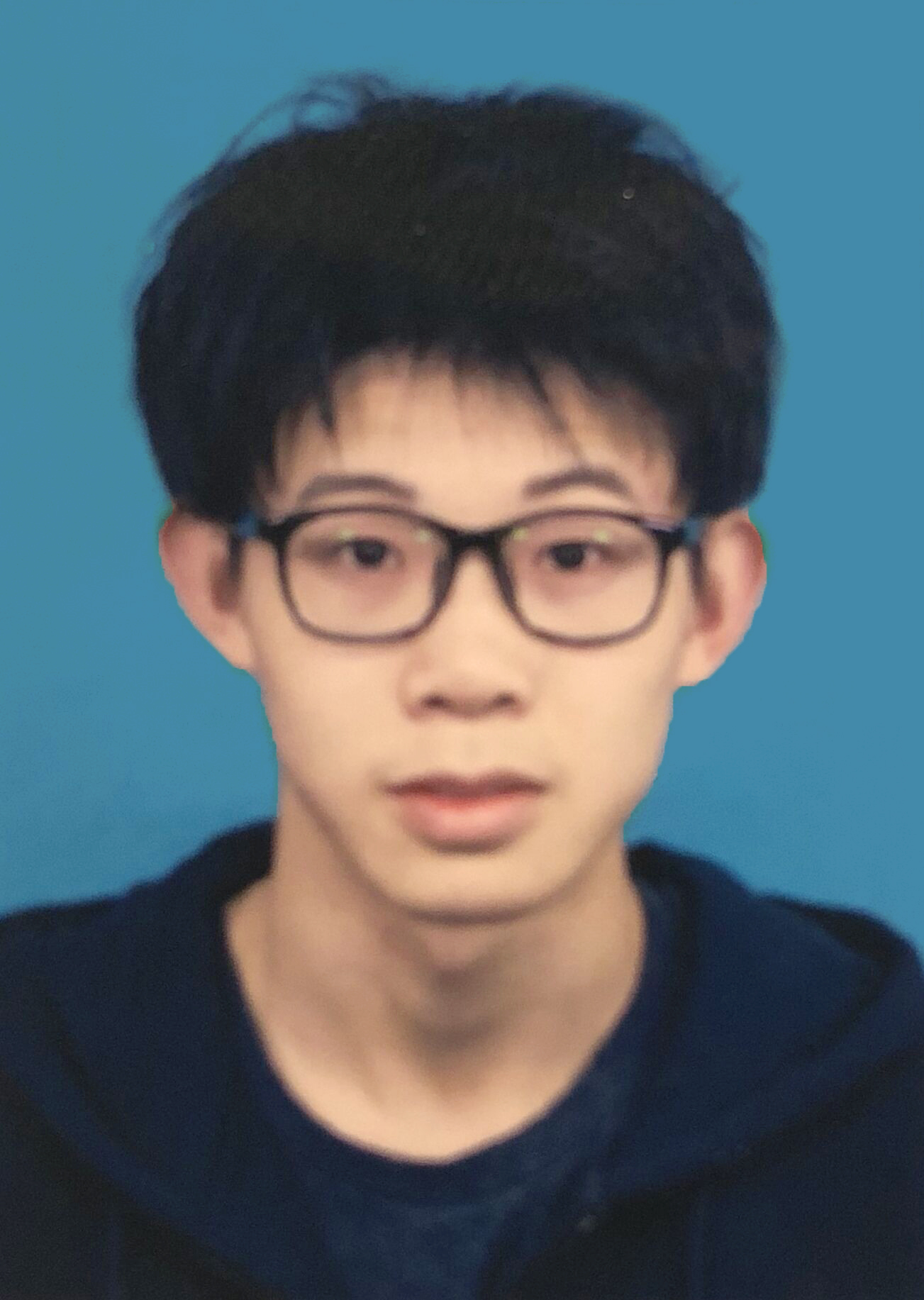}}]{Zichuan Liu}
		is currently studying toward the B.S. degree at the School of Computer Science and Technology, Wuhan University of Technology, China. His research interests include but are not limited to deep learning and data mining, with a particular focus on spatial-temporal data mining. He often participates in some kaggle competitions and won Data Science Bowl student award in 2020.
	\end{IEEEbiography}
	
	\begin{IEEEbiography}[{\includegraphics[width=1in,height=1.25in,clip,keepaspectratio]{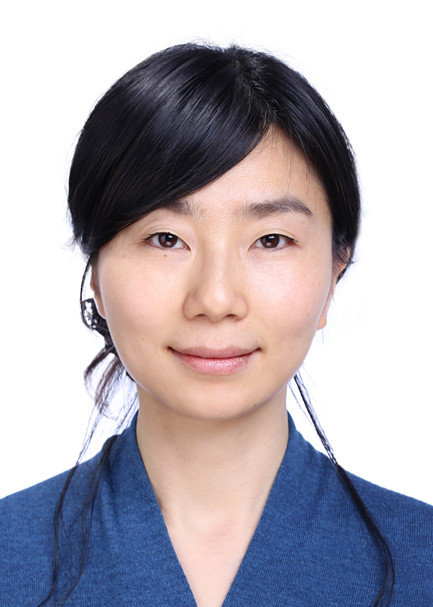}}]{Rui Zhang}
		is an Associate Professor in School of Computer Science and Technology at Wuhan University of Technology, China. She received the M.S. degree and Ph.D. degree in Computer Science from Huazhong University of Science and Technology, China. From 2013 to 2014, she was a Visiting Scholar with the College of Computing, Georgia Institute of Technology, USA. Her research interests include machine learning, network analysis and mobile computing.
	\end{IEEEbiography}
	
	
	\begin{IEEEbiography}[{\includegraphics[width=1in,height=1.25in,clip,keepaspectratio]{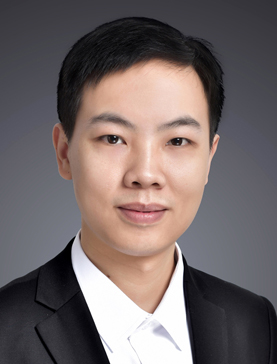}}]{Chen Wang}
		(S'10-M'13-SM'19) received the B.S. and Ph.D. degrees from the Department of Automation, Wuhan University, China, in 2008 and 2013, respectively. From 2013 to 2017, he was a postdoctoral research fellow in the Networked and Communication Systems Research Lab, Huazhong University of Science and Technology, China. Thereafter, he joined the faculty of Huazhong University of Science and Technology where he is currently an associate professor. His research interests are in the broad areas of Internet of Things, data mining and mobile computing, with a recent focus on privacy issues in wireless and mobile systems. He is a senior member of IEEE and ACM.
	\end{IEEEbiography}
	
	\begin{IEEEbiography}[{\includegraphics[width=1in,height=1.25in,clip,keepaspectratio]{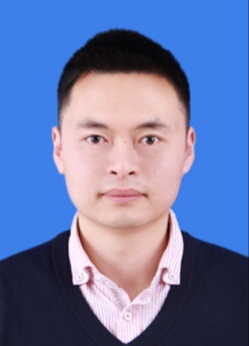}}]{Zhu Xiao}(M'15-SM'19) received the M.S. and Ph.D. degrees in communication and information systems from Xidian University, China, in 2007 and 2009, respectively. From 2010 to 2012, he was a Research Fellow with the Department of Computer Science and Technology, University of Bedfordshire, U.K. He is currently an Associate Professor with the College of Computer Science and Electronic Engineering, Hunan University, China. His research interests include wireless localization, Internet of Vehicles and intelligent transportation systems. He is a Senior Member of the IEEE. He is serving as an Associate Editor of the~\emph{IEEE Transactions on Intelligent Transportation Systems}.
	\end{IEEEbiography}

	\begin{IEEEbiography}[{\includegraphics[width=1in,height=1.25in,clip,keepaspectratio]{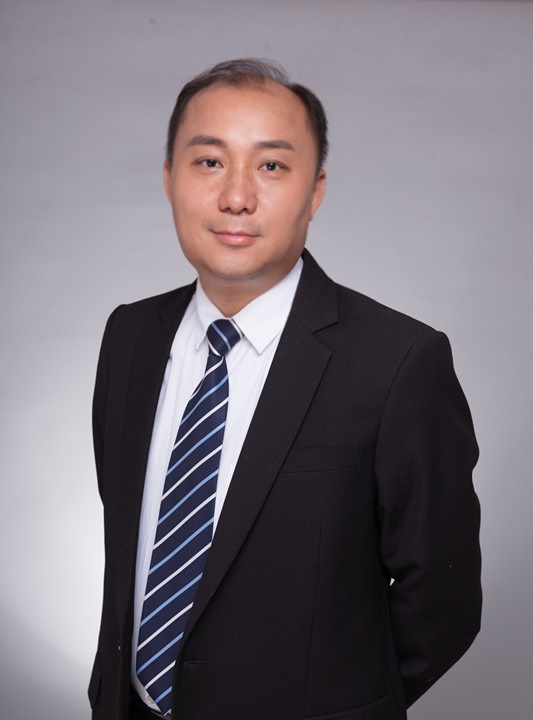}}]
		{Hongbo Jiang} (M'08-SM'15) is now a full Professor in the College of Computer Science and Electronic Engineering, Hunan University. He ever was a Professor at Huazhong University of Science and Technology. 
		He received his Ph.D. from Case Western Reserve University in 2008. His research concerns computer networking, especially algorithms and protocols for wireless and mobile networks. He is serving as the editor for IEEE/ACM Transactions on Networking, the associate editor for IEEE Transactions on Mobile Computing, ACM Transactions on Sensor Networks, IEEE Transactions on Network Science and Engineering, IEEE Transactions on Intelligent Transportation Systems, and the associate technical editor for IEEE Communications Magazine. He is a senior member of the IEEE.
		
	\end{IEEEbiography}
	
	
	

\end{document}